\documentclass[runningheads]{llncs}
\pdfoutput=1

\usepackage{times}
\usepackage{helvet}
\usepackage{courier}

\usepackage{amsmath} 

\emergencystretch=\hsize
\usepackage[english,american]{babel}

\usepackage{comment}

\usepackage{graphicx}

\usepackage{mdwlist}	

\usepackage{xspace}

\usepackage{stmaryrd}      

\long\def\symbolfootnote[#1]#2{\begingroup%
\def\thefootnote{\fnsymbol{footnote}}\footnotetext[#1]{#2}\endgroup}

\title{
Language {\fASP} with Arithmetic Expressions and Consistency-Restoring Rules
}
\author{Marcello Balduccini\inst{1} \and Michael Gelfond\inst{2}}
\institute{Kodak Research Laboratories \\
Eastman Kodak Company \\
Rochester, NY 14650-2102 USA \\
\email{marcello.balduccini@gmail.com}
\and
Computer Science Department \\
Texas Tech University \\
Lubbock, TX 79409 USA \\
\email{michael.gelfond@ttu.edu}
}

\newcommand{\tbeg}{\langle}
\newcommand{\tend}{\rangle}
\newcommand{\entails}{\models}

\newcommand{\lpnot}{\mbox{\textit{not}}\,\,}

\newcommand{\hif}{\leftarrow}

\newcommand{\rif}{\stackrel{\,\,+}{\leftarrow}}
\newcommand{\st}{\medskip\noindent}

\newcommand{\aspindent}{\hspace*{.25in}}

\newcommand{\suggestion}[2]{}

\newcommand{\T}{\mathbf{T}}
\newcommand{\F}{\mathbf{F}}

\def\beq{\begin{equation}} 
\def\eeq#1{\label{#1}\end{equation}}

\newcommand{\fASP}{ASP\{f\}}
\newcommand{\fASPcr}{ASP\{f,cr\}}
\newcommand{\claspf}{\textsc{clasp}\{f\}}
\newcommand{\clingcon}{\textsc{clingcon}}
\newcommand{\undef}{<\!\!unde\!f\!\!>}

\titlerunning{Language {\fASP} with Arithmetic Expressions and Consistency-Restoring Rules}
\authorrunning{M.~Balduccini and M.~Gelfond}

\begin{document}
\setcounter{page}{35}
\selectlanguage{american}

\maketitle

\begin{abstract}
In this paper we continue the work on our extension
of Answer Set Programming by non-Herbrand functions
and add to the language support for arithmetic expressions
and various inequality relations over non-Herbrand functions,
as well as consistency-restoring rules from CR-Prolog.
We demonstrate the use of this latest version of
the language in the representation of important
kinds of knowledge.
\end{abstract}

\section{Introduction}
In this paper we describe an extension of Answer Set Programming (ASP) 
\cite{gl91,mt99,bar03}
called {\fASPcr}. This work continues our research on
the introduction of non-Herbrand functions in ASP.

\st
In logic programming, functions are typically interpreted over
the Herbrand Universe, with each functional term $f(x)$ mapped to its
own canonical syntactical representation. That is, in most logic
programming languages, the value of an expression $f(x)$ is $f(x)$
itself, and thus, if equality is understood as identity, 
$f(x)=2$ is false.
This type of functions, the corresponding languages and efficient
implementation of solvers is the subject of a substantial amount
of research (we refer the reader to e.g. \cite{cci10,bb10}). 

\st
When representing certain kinds of knowledge, however, it is
sometimes convenient to use functions with \emph{non-Herbrand domains}
(\emph{non-Herbrand functions} for short),
i.e. functions that are interpreted over domains other
than the Herbrand Universe.
For example, when describing a domain in which people enter and
exit a room over time, it may be convenient to represent the
number of people in the room at step $s$ by means of a function
$occupancy(s)$ and to state the effect of a person entering
the room by means of a statement such as
\[
occupancy(S+1)=O+1 \hif occupancy(S)=O
\]
where $S$ is a variable ranging over the possible time steps
in the evolution of the domain and $O$ ranges over natural numbers.

\st
Of course, in most logic programming languages, 
non-Herbrand functions can still be represented, but the
corresponding encodings are not as natural and declarative
as the one above.
For instance, a common approach consists in representing the functions
of interest using relations, and then characterizing
the functional nature of these relations by writing auxiliary
axioms. In ASP, one would encode the above statement by
(1) introducing a relation $occupancy'(s,o)$, whose intuitive
meaning is that $occupancy'(s,o)$ holds iff the value of
$occupancy(s)$ is $o$;
and (2) re-writing the original statement as a rule
\begin{equation}\label{eq:herbrand}
occupancy'(S+1,O+1) \hif occupancy'(S,O).
\end{equation}
The characterization
of the relation as representing a function would be completed 
by an axiom such as
\begin{equation}\label{eq:herbrand-uniqueness}
\neg occupancy'(S,O') \hif occupancy'(S,O),\ O \not= O'.
\end{equation}
which intuitively states that $occupancy(s)$ has
a unique value.
The disadvantage of this representation is that
the functional nature of $occupancy'(s,o)$ is only stated
in (\ref{eq:herbrand-uniqueness}). When reading
(\ref{eq:herbrand}), one is given no indication that
$occupancy'(s,o)$ represents a function -- and,
before finding statements such as
(\ref{eq:herbrand-uniqueness}),
one can make no assumption about the functional
nature of the relations in a program
when a combination of (proper)
relations and non-Herbrand functions are present.
Moreover, in ASP relational encodings of functions often
pose performance issues. For example, the
grounding of rule (\ref{eq:herbrand-uniqueness}) grows
with $O(|D|^2)$ where $D$ is the range of variables
$O$ and $O'$. If $|D|$ is large, which is often the case
especially with numerical variables, the size of the grounding
can affect very negatively the overall solver performance.

\st
Various extensions of ASP with non-Herbrand functions
exist in the literature.
In \cite{cab11}, Quantified Equilibrium Logic is extended
with support for equality. 
A subset of the general language,
called FLP,
is then identified, which can be translated into normal logic
programs. Such translation makes it possible to compute
the answer sets of FLP programs using ASP solvers,
although the performance issues due to the size of the grounding remain.
\cite{lif11} proposes instead the use of second-order
theories for the definition of the semantics of
the language. 
Again, a transformation is (partially) described,
which removes non-Herbrand functions and makes it possible
to use ASP solvers for the computation of the answer sets
of programs in the extended language.
As with the previous approach, the performance issues are present.
In \cite{lw08,wyy09} the semantics is based on
the notion of reduct as in the original ASP semantics \cite{gl91}.
For the purpose of computing answer sets, 
a translation is defined, which maps programs
of the language from \cite{lw08,wyy09} to constraint satisfaction problems,
so that CSP solvers can be used for the computation
of the answer sets of programs in the extended language.
Finally, the language of {\clingcon} \cite{gos09} extends
ASP with elements from constraint satisfaction.
The {\clingcon} solver finds the answer sets of a program
by interleaving the computations of an ASP solver and of
a CSP solver.
All the approaches except for \cite{cab11} support only total functions.
While the approaches from \cite{lw08,wyy09,gos09} are computationally efficient, 
the approaches of \cite{cab11,lif11}, based on translations to ASP,
are affected by performance issues due to
the size of the grounding.
Finally, in \cite{bal12b,bal12} we
have proposed
an extension of ASP with non-Herbrand functions, called {\fASP}, 
that supports partial functions and is computationally more
efficient than \cite{cab11}.
%
%
%
%

\st
In the present paper, we extend the definition of {\fASP} from \cite{bal12b,bal12}
further,
by adding to it support for full-fledged arithmetic expressions
and for consistency restoring rules from CR-Prolog \cite{bg03}.
We also contribute our perspective to the current debate on the usefulness of
partial vs. total functions and on the role of non-Herbrand languages
in general by demonstrating the use of our extended language for
the representation of important types of knowledge and
for the encoding of some classical scenarios, pointing out the 
differences with other approaches and with ASP encodings.

\st
The rest of the paper is organized as follows.
In the next section we extend
{\fASP} with full-fledged arithmetic expressions.
In the following section we introduce consistency-restoring.
The resulting language is called {\fASPcr}.
Next, we address high-level issues of knowledge representation
in {\fASPcr} and we demonstrate
the use of {\fASPcr} for the formalization of some classical problems.
Finally, we draw conclusions and discuss
future work.

\section{{\fASP} with Arithmetic Expressions}
In this section we summarize the syntax and the
semantics of {\fASP} \cite{bal12b,bal12},
and extend the language with support for arithmetic expressions
over non-Herbrand functional terms.
For simplicity, in the rest of this paper we
drop the attribute ``non-Herbrand'' and simply
talk about functions and (functional) terms.

\st
The syntax of {\fASP} is based on 
a signature
$\Sigma=\tbeg \mathcal{C}, \mathcal{F}, \mathcal{R} \tend$
whose elements are, respectively,  finite disjoint sets of \emph{constants},
\emph{function symbols} and \emph{relation symbols}.
Some constants and function symbols are numerical (e.g. numerical constants $1$ and $5$)
and have the standard interpretation.\footnote{
In the rest of the paper, whether an element of $\Sigma$ is numerical will be clear from the context.}
A \emph{simple term} is an expression $f(c_1,\ldots,c_n)$
where $f \in \mathcal{F}$, and $c_i$'s are $0$ or more constants.
An \emph{arithmetic term} is either a simple term where $f$ is a numerical
function, or an expression constructed from such simple terms and
numerical constants using arithmetic operations, such
as $(f(c_1) + g(c_2)) / 2$ and $| f(c_1) - g(c_2) |$.
Simple terms and arithmetic terms are called \emph{terms}.
An \emph{atom} is an expression $r(c_1,\ldots,c_n)$,
where $r \in \mathcal{R}$, and $c_i$'s are constants.
The set of all simple terms that can be formed from $\Sigma$ is
denoted by $\mathcal{S}$; the set of 
all atoms from $\Sigma$ is denoted by $\mathcal{A}$.
A \emph{term-atom}, or \emph{t-atom}, is an expression of the form $f\ \texttt{op}\ g$, 
where $f$ and $g$ are terms and $\texttt{op} \in \{ =, \not=, \leq, <, >, \geq \}$.
A \emph{seed t-atom} is a t-atom of the form $f=v$ such that $f$ is a simple
term and $v$ is a constant.
All other t-atoms are called \emph{dependent}.

\st
A \emph{regular literal} is an 
atom $a$ or its strong negation $\neg a$.
A \emph{literal} is either an atom $a$, its strong negation $\neg a$,
or a t-atom.
Any literal that is not a dependent t-atom is called \emph{seed literal}.

\st
A \emph{rule} $r$ is a statement of the form:
\begin{equation}\label{eq:rule}
h \hif l_1, \ldots, l_m, \lpnot l_{m+1}, \ldots, \lpnot l_n
\end{equation}
where $h$ is a seed literal and $l_i$'s are literals.
Similarly to ASP, the informal reading of $r$ is that
a rational agent
who believes $l_1, \ldots, l_m$ and has no reason
to believe $l_{m+1}, \ldots, l_n$ must believe $h$.

\st
Given rule $r$, $head(r)$
denotes $\{ h \}$; 
$body(r)$ denotes $\{ l_1, \ldots, \lpnot l_n \}$;
$pos(r)$ denotes $\{l_1, \ldots, l_m\}$; $neg(r)$ denotes
$\{l_{m+1},\ldots,l_n\}$.

\st
A \emph{constraint} is a special type of regular rule with an empty head,
informally meaning that the condition described by
the body of the constraint must never be satisfied.
A constraint is considered a shorthand of
$\bot \hif l_1, \ldots, l_m, \lpnot l_{m+1}, \ldots, \lpnot l_n, \lpnot \bot,$
where $\bot$ is a fresh atom.

\st
A \emph{program} is a pair $\Pi=\tbeg \Sigma, P \tend$,
where $\Sigma$ is a signature and $P$ is a set of
rules. Whenever possible, in this paper
the signature is implicitly defined from the rules of
$\Pi$, and $\Pi$ is identified with its set
of rules. In that case, the signature is denoted
by $\Sigma(\Pi)$ and its elements by $\mathcal{C}(\Pi)$,
$\mathcal{F}(\Pi)$ and $\mathcal{R}(\Pi)$.
A rule $r$ is \emph{positive} if $neg(r)=\emptyset$.
A program $\Pi$ is \emph{positive} if every $r \in \Pi$ is
positive.
A program $\Pi$ is also \emph{t-atom free} if no t-atoms
occur in the rules of $\Pi$.

\st
As in ASP, variables can be used for a more compact notation.
The \emph{grounding of a rule $r$} is the set of all the
rules (its \emph{ground instances}) obtained by 
replacing every
variable of $r$ with an element of $\mathcal{C}$\footnote{
The replacement is with constants of suitable sort. We
omit the details of this process to save space.
} and by
performing any arithmetic operation over numerical constants.
For example, one of the groundings of $p(X+Y) \hif r(X),q(Y)$
is $p(5) \hif r(3),q(2)$.
The \emph{grounding of a program $\Pi$} is the set of the
groundings of the rules of $\Pi$.
A syntactic element of the language is \emph{ground} if it 
contains neither variables nor arithmetic operations over 
numerical constants and \emph{non-ground} otherwise.
For example, $p(5)$ is ground while $p(X+Y)$ and $p(3+2)$ are non-ground.

\st
The semantics of a non-ground
program is defined to coincide with the semantics of its grounding.
The semantics of ground {\fASP} programs is defined
below. 
It is worth noting that the semantics of {\fASP}
is obtained from that of ASP in \cite{gl91} by simply extending
entailment to t-atoms.

\st
In the rest of this section, we consider only ground
terms, literals, rules and programs and thus omit the word ``ground.''
A set $S$ of seed literals is \emph{consistent} if
(1) for every atom $a \in \mathcal{A}$, $\{ a, \neg a \} \not\subseteq S$;
(2) for every term $t \in \mathcal{S}$ and $v_1$, $v_2 \in \mathcal{C}$ such that
$v_1 \not= v_2$, $\{ t = v_1, t = v_2 \} \not\subseteq S$.
Hence, $S_1=\{ p, \neg q, f=3 \}$ and
$S_2=\{ q, f=3, g=2 \}$ are consistent, while
$\{ p, \neg p, f=3 \}$ and $\{ q, f=3, f=2 \}$ are not.

\st
The \emph{value} of a simple term $t$ w.r.t. a consistent set $S$ 
of seed literals (denoted by
$val_S(t)$) is $v$ iff $t = v \in S$.
If, for every $v \in \mathcal{C}$, $t=v \not\in S$,
the value of $t$ w.r.t. $S$ is \emph{undefined}.
The value of a arithmetic term $t$ w.r.t. $S$
is obtained by applying the usual rules of arithmetic
to the values of the terms in $t$ w.r.t. $S$,
if the values of all the terms in $t$ are defined; otherwise
its value is undefined.\footnote{This definition
does not adequately capture the value of expressions such as
$0*f$ in the presence of undefined terms.
We plan to address this and some related issues in a later paper.}
Finally, the value of a constant $v \in \mathcal{C}$ w.r.t.
$S$ ($val_S(v)$) is $v$ itself.
For example given $S_1$ and $S_2$ as above, 
$val_{S_2}(f)$ is $3$ and $val_{S_2}(g)$ is $2$,
whereas $val_{S_1}(g)$ is undefined. Given
$S_1$ and a signature with $\mathcal{C}=\{ 0, 1 \}$,
$val_{S_1}(1)=1$.

\st
A literal $l$ is \emph{satisfied} by a consistent set $S$
of seed literals under the following conditions:
(1) if $l$ is a seed literal, then $l$ is satisfied by $S$
iff $l \in S$; (2) if $l$ is a dependent t-atom 
of the form $f\ \texttt{op}\ g$, then $l$ is  \emph{satisfied} by
$S$ iff both $val_S(f)$ and $val_S(g)$ are defined, 
and they satisfy the equality or inequality relation $\texttt{op}$
according to the usual arithmetic interpretation.
Thus, seed literals $q$ and $f=3$ are satisfied by $S_2$;
$f \not= g$ is also satisfied by $S_2$ because $val_{S_2}(f)$ and
$val_{S_2}(g)$ are defined, and $val_{S_2}(f)$ is different from $val_{S_2}(g)$.
Conversely, $f = g$ is not satisfied, because
$val_{S_2}(f)$ is different from $val_{S_2}(g)$.
The t-atom $f \not= h$ is also not satisfied by $S_2$, because
$val_{S_2}(h)$ is undefined.
When a literal $l$ is satisfied (resp., not satisfied) by
$S$, we write $S \entails l$ (resp., $S \not\entails l$).

\st
An \emph{extended literal} is a literal $l$ or an expression
of the form $\lpnot l$.
An extended literal $\lpnot l$ is satisfied by a consistent set $S$
of seed literals ($S \entails \lpnot l$) if
$S \not\entails l$. Similarly, $S \not\entails \lpnot l$
if $S \entails l$.
Considering set $S_2$ again, extended literal $\lpnot f=h$
is satisfied by $S_2$, because $f=h$ is not satisfied by $S_2$.

\st
Finally, a set $E$ of extended literals is satisfied by 
a consistent set $S$ of seed literals ($S \entails E$) if
$S \entails e$ for every $e \in E$.

\st
Next, we define the semantics of {\fASP}.
A set $S$ of seed literals is \emph{closed} under positive rule $r$ if
$S \entails h$, where $head(r)=\{ h \}$, whenever $S \entails pos(r)$.
Hence, set $S_2$ described earlier is closed under
$f = 3 \hif g \not= 1$ and (trivially) under $f = 2 \hif r$,
but it is not closed under $p \hif f = 3$, because $S_2 \entails f = 3$
but $S_2 \not\entails p$.
$S$ is closed under $\Pi$
if it is closed under every rule $r \in \Pi$.

\begin{definition}
A set $S$ of seed literals is an \emph{answer set} 
of a positive program $\Pi$
if it is consistent and closed under $\Pi$, and is
minimal (w.r.t. set-theoretic inclusion) among the sets of
seed literals that satisfy such conditions.
\end{definition}
Thus, the program
$\{ p \hif f=2.\ \ 
f=2.\ \ 
q \hif q. \}$
has one answer sets, $\{ f = 2, p \}$.
The set $\{ f = 2 \}$ is not closed under the
first rule of the program, and therefore is not
an answer set. The set $\{ f = 2, p, q \}$
is also not an answer set, because it is not minimal (it is
a proper superset of another answer set).
Notice that positive programs may have no answer set.
For example, the program
$\{ f=3.\ \ 
f=2 \hif q.\ \ q. \} $
has no answer set. Programs that have answer sets
(resp., no answer sets) are called \emph{consistent}
(resp., \emph{inconsistent}).

\st
Positive programs 
enjoy the following property:
\begin{proposition}\label{prop:uniqueness}
Every consistent positive {\fASP} program $\Pi$
has a unique answer set.
\end{proposition}

\st
Next, we define the semantics of arbitrary {\fASP}
programs.

\begin{definition}
The \emph{reduct} of a program $\Pi$ w.r.t.
a consistent set $S$ of seed literals is the set $\Pi^S$ consisting
of a rule $head(r) \hif pos(r)$ (the \emph{reduct} of $r$ w.r.t. $S$)
for each rule $r \in \Pi$ 
for which $S \entails body(r) \setminus pos(r)$.
\end{definition}
\begin{example}
Consider a set of seed literals
$S_3=\{ g=3, f=2, p, q \}$, and program $\Pi_1$:
\[
\begin{array}{l}
r_1: p \hif f=2, \lpnot g = 1, \lpnot h = 0. \\
r_2: q \hif p, \lpnot g \not= 2. \\
r_3: g=3. \\
r_4: f=2.
\end{array}
\]
and let us compute its reduct. For $r_1$, first
we have to check if $S_3 \entails body(r_1) \setminus pos(r_1)$,
that is if $S_3 \entails \lpnot g = 1, \lpnot h = 0$.
Extended literal $\lpnot g = 1$ is satisfied by $S_3$ only 
if $S_3 \not\entails g = 1$.
Because $g = 1$ is a seed literal, it is satisfied by $S_3$ if
$g = 1 \in S_3$. Since $g = 1 \not\in S_3$, we conclude
that $S_3 \not\entails g = 1$ and thus $\lpnot g = 1$ is 
satisfied by $S_3$. In a similar way, we conclude that
$S_3 \entails \lpnot h = 0$. Hence,
$S_3 \entails body(r_1) \setminus pos(r_1)$. Therefore,
the reduct of $r_1$ is $p \hif f = 2$.
For the reduct of $r_2$, notice that $\lpnot g \not= 2$ is not
satisfied by $S_3$. In fact, $S_3 \entails \lpnot g \not= 2$
only if $S_3 \not\entails g \not= 2$.
However, 
it is not difficult to show that 
$S_3 \entails g \not= 2$:
in fact,
$val_{S_3}(g)$ is defined and $val_{S_3}(g) \not= 2$.
Therefore, $\lpnot g \not= 2$ is not satisfied by $S_3$,
and thus the reduct of $\Pi_1$ contains no rule for $r_2$.
The reducts of $r_3$ and $r_4$ are the rules themselves.
Summing up, $\Pi_1^{S_3}$ is
$\{ r_1': p \hif f=2,\ 
r_3': g=3, \ 
r_4': f=2 \}$
\end{example}
The semantics of arbitrary {\fASP} programs is given by the
following definition:
\begin{definition}
Finally, a consistent set $S$ of seed literals is an \emph{answer set}
of $\Pi$ if $S$ is the answer set of $\Pi^S$.
\end{definition}

\begin{example}
By applying the definitions given earlier, it is not difficult to
show that an answer set of $\Pi_1^{S_3}$ is
$\{ f = 2, g = 3, p \} = S_3$. Hence, $S_3$ is an answer set of
$\Pi_1^{S_3}$. Consider instead $S_4 = S_3 \cup \{ h = 1 \}$.
Clearly $\Pi_1^{S_4} = \Pi_1^{S_3}$.
From the uniqueness of the answer sets of positive 
programs, it follows immediately
that $S_4$ is not an answer set of $\Pi_1^{S_4}$. Therefore, 
$S_4$ is not an answer set of $\Pi_1$.
\end{example}

\section{{\fASPcr}: Consistency-Restoring Rules in {\fASP}}
In this section we extend {\fASP} by consistency-restoring
rules from CR-Prolog \cite{bg03}.
We denote the extended language by {\fASPcr}.
As discussed in the literature on CR-Prolog, consistency-restoring
rules are convenient for the formalization of various types of 
knowledge and of reasoning tasks. Later in this paper
we show how consistency-restoring rules are useful for the
formalization of knowledge about non-Herbrand functions as well.

\st
A \emph{consistency-restoring rule} (or \emph{cr-rule}) is a statement
of the form:
\begin{equation}\label{eq:cr-rule}
h \rif l_1, \ldots, l_m, \lpnot l_{m+1}, \ldots, \lpnot l_n.
\end{equation}
where $h$ is a seed literal and $l_i$'s are literals.
The intuitive reading of the statement is that a reasoner
who believes $\{ l_1, \ldots, l_m \}$ and has no reason to believe
$\{l_{m+1}, \ldots, l_n\}$, \emph{may possibly} believe $h$.
The implicit assumption is that this possibility 
is used as little as possible, only when the reasoner cannot
otherwise form a non-contradictory set of beliefs. 

\st
By \emph{{\fASPcr} program} we mean a pair $\tbeg \Sigma, \Pi \tend$,
where $\Sigma$ is a signature and $\Pi$ is a set of rules and
cr-rules over $\Sigma$.

\st
Given an {\fASPcr} program $\Pi$, we denote the set of
its rules by $\Pi^r$ and the set of its cr-rules
by $\Pi^{cr}$. By $\alpha(r)$ we denote the 
rule obtained from cr-rule $r$ by replacing symbol
$\rif$ with $\hif$. Given a set of cr-rules $R$,
$\alpha(R)$ denotes the set obtained by applying
$\alpha$ to each cr-rule in $R$.
The semantics of {\fASPcr} programs is defined in
two steps.

\begin{definition}[Answer Sets of CR-Rule Free Programs]\label{def:as-cr-rule-free}
The \emph{answer sets} of a cr-rule free {\fASPcr} program are
the answer sets of the corresponding {\fASP} program.
\end{definition}

\begin{definition}\label{def:abductive-support}
Given an arbitrary {\fASPcr} program $\Pi$, a subset $R$ of $\Pi_{cr}$ is
an \emph{abductive support} of $\Pi$ if
$\Pi^r \cup \alpha(R)$ is consistent and $R$ is
set-theoretically minimal among the
sets satisfying this property.
\end{definition}

\begin{definition}[Answer Sets of Arbitrary Programs]\label{def:as-arbitrary-cr}
For an arbitrary {\fASPcr} program $\Pi$, a set of literals $A$ is an
\emph{answer set} of $\Pi$ if it is an answer set of the
program $\Pi^r \cup \alpha(R)$ for some abductive support
$R$ of $\Pi$.
\end{definition}
Although out of the scope of the present paper, 
it is also possible to extend 
{\fASPcr} to allow for the specification of CR-Prolog-style
preferences over cr-rules.

\section{Knowledge Representation with {\fASPcr}}
In this section we demonstrate the use of {\fASPcr} for
the formalization of certain types of knowledge.
Whenever appropriate, we also compare with ASP and with other 
extensions of ASP by non-Herbrand functions.

\st
Consider a scenario in which data from a town registry 
about births and deaths is
used to determine who should receive a certain tax bill.
The registry lists the year of birth and the year of death
of a person. If a person is alive, no year of death is in
the registry. 
The tax bill should only be sent to living people
who are between 18 and 65 years old.
To ensure robustness, we want to be able to
deal with (infrequently) missing information.
Hence, whenever there is uncertainty (represented by an atom $uncertain(p)$)
about whether a person should receive the tax bill or not, 
a manual check will be performed.
The first requirement can be encoded in {\fASPcr} by the rule:
\[
\begin{array}{l}
bill(P) \hif \\
\aspindent person(P), \\
\aspindent age(P) \geq 18, age(P) \leq 65, \\
\aspindent \lpnot uncertain(P), \\
\aspindent \lpnot \neg bill(P).
\end{array}
\]
Relation $person$ defines a list of people known to the system.
To shorten the rules, from now on we will
implicitly assume the occurrence of an atom $person(P)$ in
every rule where $P$ occurs.
The condition $\lpnot \neg bill(P)$ allows one to specify
exceptions in the usual way. For example, the tax might be waived
for low-income people:
\[
\neg bill(P) \hif low\_income(P).
\]
Similarly, condition $\lpnot uncertain(P)$ ensures that the bill is not sent
if there is uncertainty about whether the person is subject
to the tax.

\st
Next, we define a person's current age based on
their year of birth. Following
intuition, we define $age$ only for people who are alive.
\begin{align}
&has\_birth\_year(P) \hif birth\_year(P)=X.\label{eq:age-1} &\\
&\neg has\_birth\_year(P) \hif \lpnot has\_birth\_year(P).\label{eq:age-1b} &\\
\nonumber \\
&has\_death\_year(P) \hif death\_year(P)=X.\label{eq:age-2} &\\
&\neg has\_death\_year(P) \hif \lpnot has\_death\_year(P).\label{eq:age-2b} &\\
\nonumber \\
&\neg alive(P) \hif has\_death\_year(P).\label{eq:age-3} &\\
&alive(P) \hif has\_birth\_year(P), \neg has\_death\_year(P).\label{eq:age-4} &\\
\nonumber \\
&age(P)=X \hif alive(P), X=current\_year - birth\_year(P).\label{eq:age-5} &
\end{align}
Rules (\ref{eq:age-1}) and (\ref{eq:age-2}) determine
when information about a person's birth and death year is
available. Rules (\ref{eq:age-1b}) and (\ref{eq:age-2b})
formalize the closed world assumption (CWA) of relations $has\_birth\_year$
and $has\_death\_year$.
Although this encoding of CWA is common practice in ASP,
it plays an important role in the distinction between
languages with partial functions and languages with total functions,
as we discuss later.
Rule (\ref{eq:age-3}) states that it is possible to conclude that
a person is dead if a year of death is found in the registry.
Rule (\ref{eq:age-4}) states that a person is alive if the
registry contains a year of birth and does not contain information
about the person's death. 
Finally, rule (\ref{eq:age-5}) calculates a person's age. $current\_year$
is a function of arity $0$ whose value corresponds to the current year.

\st
The next set of rules deals with the possibility of information 
missing from the registry. One important case to consider
is that in which information about a person's death is accidentally missing
from the registry. In (\ref{eq:age-4}) a person is assumed
to be alive unless evidence exists about the person's death.
This modeling choice is
justified because missing information is assumed to be infrequent.
Exceptional conditions can be dealt with by requesting a manual
check on whether the person should receive the tax bill.
Rule (\ref{eq:uncertain-1}) below
states that one such case is when a person's age
according to the registry is beyond that person's maximum
life span.
\begin{align}
&uncertain(P) \hif alive(P), age(P) \geq max\_span(P). \label{eq:uncertain-1}&\\
&max\_span(P)=92 \hif \lpnot max\_span(P) \not= 92. \label{eq:uncertain-2}& \\
&max\_span(P)=100 \hif long\_lived\_family(P). \label{eq:uncertain-2b}& \\
\nonumber \\
&uncertain(P) \hif \lpnot alive(P), \lpnot \neg alive(P). \label{eq:uncertain-3}&
\end{align}
Rule (\ref{eq:uncertain-2}) provides a simple definition
of a person's maximum life span, stating that, normally, a person's
maximum life span is $92$ years. Note that the rule is written
in the form of a \emph{default over non-Herbrand functions}. This
makes it possible for example to predict a different
life span depending on a person's medical or family history.
Along the lines of \cite{bg94},
the reading of (\ref{eq:uncertain-2}) is 
``if $P$'s maximum life span \emph{may be} $92$, then
it is $92$.''
Generally speaking, 
an expressions of the form $\lpnot f \not= g$ can be viewed to intuitively
state ``$f$ \emph{may} be (equal to) $g$''.
Rule (\ref{eq:uncertain-2b}) encodes one possible exception to
the default, for people
from families with a history of long life spans.
Rule (\ref{eq:uncertain-3}) covers instead the case
in which the system couldn't determine if a person is dead or
alive. The formalization of this type of test has already been
covered in the literature
and is shown here for completeness, and using a 
a slightly simplified encoding.
A discussion on this topic and proposals for more sophisticated
formalizations can be found in \cite{gel91b,gel11}.

\st
It is currently a source of debate \cite{lif11,cab11} whether
support for partial functions should be allowed in languages with
non-Herbrand functions.
Although of course from the point of view of computational complexity
partial and total functions in this context are equivalent,
we believe that the following elaboration of the tax-bill scenario
appears to show that the availability of partial functions is indeed 
important.

\st
First of all, notice that, in a language that only supported
total functions, the scenario discussed so far would have to be
formalized by introducing a special constant.
This special constant is to be
used when the birth or death years are unknown. For the
sake of this discussion, let us denote the special constant by
${\undef}$.

\st
Notice that, to allow for the use of ${\undef}$, a design requirement 
would have to be imposed on the town registry so that 
entries that do not have a value are set to ${\undef}$.
It is worth mentioning that one might be
tempted to avoid the use
of ${\undef}$ and instead reason by cases, one for each
possible value of the year of birth or death. This
approach however
does not appear to work well in this scenario.
In fact, in this scenario it is important to know
whether the year of death is present in the registry at all -- see
e.g. rules (\ref{eq:age-1}) and (\ref{eq:age-2}). When
reasoning by cases, it is not possible to reason
about this circumstance from within the program, 
unless rather sophisticated extensions of ASP 
such as \cite{gel91b,gel11} are used.

\st
Going back to our scenario, suppose that we want to incorporate in
our system information from a database
of deadly car accidents.
Suppose the database consists of statements
of the form $died(p,y)$, where $p$ is a person and
$y$ is the year in which the deadly accident occurred.
If there are inconsistencies between the
town registry and the accident database,
we would like to give precedence to the former.
This can be easily formalized in {\fASPcr}
with:
\begin{align}
&death\_year(P)=Y \hif died(P,Y), \lpnot death\_year(P) \not= Y. \label{eq:accident-1}&
\end{align}
Informally, the rule states that, if $p$ is reported to have
died in a car accident in year $y$, then that is assumed to
be $p$'s death year, unless the town registry contains
information to the contrary.

\st
It is important to notice that our {\fASPcr} formalization
makes it possible to incorporate the car accident database
in a completely incremental fashion. No
changes are needed to the rules we showed earlier.
This is possible mainly because $death\_year$ is a partial
function.

\st
Let us see now how using a language with total functions 
would affect the incorporation of the car accident database.
Let us consider a situation in which no death year
is specified for person $p$ in the town registry,
but an entry $died(p,1998)$ exists in the car accident database.
As discussed above, in a language that only supported total functions,
$death\_year(p)$ would have to be set to ${\undef}$ in the
town registry. Hence, the body of rule (\ref{eq:accident-1}) would
never be satisfied.

\st
To the best of our knowledge, working around
this issue when using a language with total functions
is non-trivial and the solutions are characterized by
reduced elaboration tolerance.
For example, one possible method consists in introducing a relation
$determined\_death\_year(P)$ that encodes
the \emph{overall} belief of the system about a person's
death year.
By default, the value of $determined\_death\_year(P)$
is obtained from the town registry. When
that value is undefined, a death year can be derived from 
the car accident database by means of a rule similar to
(to avoid using a specific language from the
literature, we write the rule in the syntax of {\fASPcr}):
\[
\begin{array}{l}
determined\_death\_year(P)=Y \hif \\
\aspindent died(P,Y), \\
\aspindent \lpnot determined\_death\_year(P) \not= Y.
\end{array}
\]
Furthermore, any rule that previously involved relation $death\_year$
would have to be modified to use the new relation.
This process, although seemingly harmless on the surface,
tends to be error-prone and the corresponding encoding is hardly elaboration tolerant. 
Every time information from another database had to be incorporated,
more changes to the existing program are required --
for example, the reader may want to consider would happen if 
one had to incorporate  
information about births from e.g. a health insurance database.

\st
Up to this point we have discussed cases in which 
the use of total functions appears to cause some issues.
One may be wondering whether it is possible to represent
total functions in {\fASPcr}, and if there are any
drawbacks.

\st
To address this topic, let us 
suppose that we would like to determine the number of dependents
of a person. This information could be used for example
in order to ensure that certain individuals are waived
from paying the tax discussed earlier. For simplicity of
presentation, however, we discuss the determination of
the number of dependents independently of the code shown earlier.

\st
Let us assume that the number of a person's dependents 
is found in their tax return, if one exists.
If no tax return has been filed, we would like to 
consider separately each possible case, corresponding to a number
of dependents ranging between $0$ and $max_d$.
The number of dependents can then be viewed as 
a total function.

\st
In our formalization,
an atom of the form $return\_dep(p,d)$ states that $p$
has $d$ dependents according to $p$'s latest tax return (or, equivalently,
a function could be used instead of a relation).
We will represent the number of a person $p$'s dependents
by means of a function $dependents(p)$.

\st
A straightforward formalization, $\Pi^i_1$, of such a total 
function is:
\begin{align}
&dependents(P)=D \hif return\_dep(P,D). \label{eq:pi^i_1-1}&\\
&dependents(P)=V \hif \lpnot dependents(P) \not= V. \label{eq:pi^i_1-2}&
\end{align}
Rule (\ref{eq:pi^i_1-1}) states that the number of
dependents can be obtained from the tax return, if
available.
Rule (\ref{eq:pi^i_1-2}) states that a person can have
any number of dependents, unless there is reason to
believe otherwise. Here and below we assume that variable $V$ ranges
over the domain $[0,max_d]$ (this can be easily implemented
by adding a condition $dom(V)$ and a suitable definition
of relation $dom$).

\st
$\Pi^i_1$ formalizes the nature of
total function $dependents$ for simple situations.
However, suppose that one wanted to take into account
the case in which $p$'s tax return was audited and
the number of $p$'s dependents found to be different from
what was stated in the tax return. Rule (\ref{eq:pi^i_1-1})
does not properly deal with this case, because it prevents
one from overriding a person's dependents based on information
from the audit.
So, a different formalization of total functions is
needed to deal with more sophisticated examples.

\st
One might then be tempted to rewrite (\ref{eq:pi^i_1-1}) as a default
and to add a suitable exception, obtaining $\Pi^i_2$:
\begin{align}
&dependents(P)=D \hif return\_dep(P,D), \lpnot dependents(P) \not= D. \label{eq:pi^i_2-1} &\\
&dependents(P)=V \hif \lpnot dependents(P) \not= V. \label{eq:pi^i_2-2}&\\
&dependents(P)=D \hif assessed\_deps(P,D).\label{eq:pi^i_2-3} &
\end{align}
Unfortunately this formalization does not yield the intended
answers because of the interaction between defaults (\ref{eq:pi^i_2-1}) 
and (\ref{eq:pi^i_2-2}):
consider a person $p_1$ with $3$ dependents according to their tax return,
$I_1=\{ return\_deps(P,3) \}$. One might expect $I_1 \cup \Pi^i_2$
to yield the conclusion $dependents(p)=3$, but in fact the program has
multiple answer sets, enumerating all the possible numbers of dependents
between $0$ and $max_d$.
This is an instance of a phenomenon already 
studied in the literature (see e.g. \cite{gelfond-3}),
which can be circumvented by properly prioritizing the defaults of $\Pi^i_2$.
Doing so however tends to affect the elaboration
tolerance of the encoding (e.g. in case further defaults must
be added) and is somewhat cumbersome and error-prone.

\st
A more robust and elaboration tolerant approach 
relies on the use of cr-rules. 
Intuitively, in this approach, a cr-rule determines when
to trigger the default behavior of considering all
the possible values of a total function.
Consider the following program, $\Pi^i_3$:
\begin{align}
&dependents(P)=D \hif return\_deps(P,D), \lpnot dependents(P) \not= D. \label{eq:pi^i_3-former-1} &\\
&dependents(P)=D \hif assessed\_deps(P,D).\label{eq:pi^i_3-former-3} &\\
\nonumber \\
&has\_dep\_info(P) \hif dependents(P)=D. \label{eq:pi^i_3-1} &\\
&\hif \lpnot has\_dep\_info(P). \label{eq:pi^i_3-2} &\\
&dependents(P)=D \rif. \label{eq:pi^i_3-3}&
\end{align}
Program $\Pi^i_3$ is obtained from $\Pi^i_2$ by replacing rule
(\ref{eq:pi^i_2-2}) by (\ref{eq:pi^i_3-1}-\ref{eq:pi^i_3-3}).
Rules (\ref{eq:pi^i_3-1}-\ref{eq:pi^i_3-2}) intuitively
state that the number of dependents must be known for every person.
Cr-rule (\ref{eq:pi^i_3-3}) intuitively states that 
it is possible to assume that a person has any number of dependents,
but that this possibility
should be used only if strictly necessary and in order to restore
consistency.

\st
It is not difficult to see that $\Pi^i_3$ yields the expected
conclusions. To this extent, it is important to notice 
that cr-rule (\ref{eq:pi^i_3-3})
will only be used for people for whom no
other dependent information is available.
In fact, let $I_3$ be a set of facts providing partial information
about the dependents of a group of people, and $U_3=\{p_1, \ldots, p_u\}$ be 
the set of people from $I_3$ for whom no dependent information is available.
According to Definition \ref{def:abductive-support}, any abductive support 
of $\Pi^i_3 \cup I_3$ must
contain, for every $p_i \in U_3$, a ground cr-rule 
$dependents(p_i)=d \rif$ for some $d$. Let now $R_3$ be the set of all such cr-rules,
and consider a person $p'$ for whom dependent information is provided in $I_3$.
The corresponding set $R_3'=R_3 \cup \{ dependents(p')=d' \rif \}$ is not an 
abductive support of $\Pi^i_3 \cup I_3$ because it is
not set-theoretically minimal. Hence, cr-rule (\ref{eq:pi^i_3-3})
is only used for the people in $U_3$.

\st
It is not difficult to see that this approach for the encoding 
of total functions in {\fASPcr} is applicable in general, and
that (\ref{eq:pi^i_3-1}) can be rewritten
as a general, domain-independent axiom (an example of a
domain-independent axiom can be found in the next section).

\section{Some Modeling and Solving Tasks in {\fASPcr}}
In this section we demonstrate the use of {\fASPcr}
for a sample of modeling and solving tasks from the literature.
We also include a (partial) discussion of the features of our
encodings in relation with other methods for representing
non-Herbrand functions.
We refer the reader to the description and original encodings
from {http://www.cs.uni-potsdam.de/$\sim$torsten/kr12tutorial}.

\st
\emph{Water Buckets on a Scale (page 216).} In this scenario,
one bucket is placed on each arm of a two-armed scale.
Each bucket initially contains an amount of water between $0$ and
$100$. All amounts of water in this scenario are represented by
integer values.
At every time step, an agent must pour an amount $k$, $1 \leq k \leq max_w$
of water into one of the buckets.\footnote{We deviate slightly
from the original scenario in that the agent is allowed to decide
how much water is to be poured.}
The agent's goal is to balance the two buckets on the scale.
The {\fASPcr} encoding, $\Pi^w$, of this scenario is:
\small
\begin{align}
&bucket(a).\ \ bucket(b). \nonumber &\\
&1 \{ pour(B,T,K) : bucket(B) : K \geq 1 : K \leq max_w \} 1 \hif time(T), T < t. &\label{eq:w-1}\\
&poured(B,T)=K \hif pour(B,T,K).&\label{eq:w-1b}\\
&volume(B,T+1)=V \hif V=volume(V,T)+poured(B,T).&\label{eq:w-2}\\
&volume(B,T+1)=volume(B,T) \hif \lpnot volume(B,T+1) \not= volume(B,T).&\label{eq:w-3}\\
&heavier(B,T) \hif bucket(B), bucket(C), time(T), volume(C,T) < volume(B,T).&\label{eq:w-4}\\
&\hif bucket(B), heavier(B,t).&\label{eq:w-5}
\end{align}
\normalsize
Rule (\ref{eq:w-1}) states that the agent can pour any allowed
amount of water in any one bucket at every time step. For
compactness, the rule uses the syntax of choice rules from
\textsc{lparse} and \textsc{gringo}. Extending the definition
of {\fASPcr} to support choice rules is trivial.
Rule (\ref{eq:w-1b}) states that the amount of water poured
as a consequence of action $pour(b,t,k)$ is $k$.
Rule (\ref{eq:w-2}) encodes a dynamic law;
it states that when water is poured into a bucket,
the volume of water in the bucket increases by
the amount of water poured. We assume that a suitable domain
has been specified for variable $V$.
Rule (\ref{eq:w-3}) formalizes the inertia axiom. It
states that the volume of water in a bucket stays the
same unless it is forced to change.
Rule (\ref{eq:w-4}) describes the conditions under which
a buckets is heavier than the other.
Finally, rule (\ref{eq:w-5}) states that it is impossible
for a bucket to be heavier than the other at the end of
the execution of the plan.

\st
It is worth observing that, as prescribed by good knowledge
representation principles, in $\Pi^w$ the inertia
axiom is written without 
references to the occurrence of any action.
This allows for a rather elaboration tolerant encoding.
In the original \textsc{clingcon} encoding, on the other hand,
the inertia axiom mentions the occurrence of actions:
\small
\begin{align}
&amount(B,T) \$== 0 \hif \lpnot pour(B,T), bucket(B), T<t.\label{eq:clingcon^w-1}&\\
&volume(B,T+1) \$== volume(B,T)+amount(B,T) \hif bucket(B), T<t.&
\end{align}
\normalsize
This encoding is arguably less elaboration tolerant than $\Pi^w$:
for example, the \textsc{clingcon} inertia axiom (\ref{eq:clingcon^w-1}) must be modified
whenever new actions are introduced in the representation, 
while the inertia axiom from $\Pi_w$ does not have to be changed,
and the whole program can be extended in a completely incremental fashion.
In fact, {\fASPcr} makes it possible to encode the inertia 
axiom in a form that is even more general than that of rule (\ref{eq:w-3}):\footnote{
To complete the encoding (\ref{eq:w-2}) and (\ref{eq:w-4}) have to be modified 
in a straightforward way to use $val(\cdot,\cdot)$
as well. From a technical perspective, in this encoding ground expressions
$volume(a)$ and $volume(b)$ are viewed as constants. This is possible because
no assumptions are made about the set of constants in our definition
of the language. Alternatively, one could of course extend the language
with Herbrand function symbols and of Herbrand terms, at the cost of a slightly
more complex presentation.
}
\small
\begin{align}
&num\_fluent(volume(B)) \hif bucket(B).&\label{eq:w2-1}\\
&val(N,T+1)=val(N,T) \hif num\_fluent(N),\lpnot val(N,T+1) \not= val(N,T).&\label{eq:w2-2}
\end{align}
\normalsize
Rule (\ref{eq:w2-1}) states that $volume(\cdot)$ is a ``numerical fluent''.
Rule (\ref{eq:w2-2}) states that the value of any numerical fluent remains the
same over time unless it is forced to change.
The advantage of this generalized form of the inertia axiom 
is that the corresponding rules apply without changes to any numerical fluent, 
so that now the addition of new numerical fluents to the encoding can be fully
incremental as well.

\st
From the point of view of the size of the grounding, the
\textsc{clingcon} encoding is however superior to $\Pi^w$, because
in $\Pi^w$ rule (\ref{eq:w-2}) must be grounded for every 
possible value of variable $V$, while in the \textsc{clingcon}
encoding the grounding is entirely independent of the
volume of water in the buckets.
On the other hand, the size of the grounding of $\Pi^w$
is substantially better than the best ASP encodings
that we are aware of. In the ASP encodings, in fact,
the grounding of the inertia axiom grows proportionally to
the \emph{square} of the domain of variable $V$.
A similar phenomenon can be observed in the encodings
based on the languages of \cite{cab11,lif11}, since 
in those approaches computation of the answer sets 
is performed by translating the programs to ASP.

\st
\emph{N-Queens (page 176)}. In this
scenario an agent must place $n$ queens
on an $n \times n$ chess board so that no queen can
attack another. In this scenario
the size of the grounding and the execution time tend to grow
quickly with the increase of parameter $n$. In straightforward
ASP encodings, the growth
of the grounding is due to the tests ensuring
that no queen can attack another. $\Pi^q_1$ shows
one possible ASP encoding:
\[
\small
\begin{array}{l}
\hif queen(X_1,Y_1), queen(X_1,Y_2), Y_1 < Y_2.\\
\hif queen(X_1,Y_1), queen(X_2,Y_1), X_1 < X_2.\\
\hif queen(X_1,Y_1), queen(X_2,Y_2), X_1 < X_2, X_2 - X_1 = |Y_2 - Y_1|.
\end{array}
\normalsize
\]
Conditions $Y_1 < Y_2$ and $X_1 < X_2$ are introduced
in order to break symmetries.
The last rule is the most problematic with respect to
the size of the grounding, because its grounding
grows roughly with $O(n^4)$. Several modifications of $\Pi^q_1$ are
known, which decrease the size of the grounding.\footnote{
See especially {http://www.cs.uni-potsdam.de/$\sim$torsten/kr12tutorial}.
}
However, it is often argued that these modifications
make the corresponding encodings either less declarative, or
less elaboration tolerant. Certainly, most of the modifications
achieve performance by a less straightforward encoding
of the constraints of the problem.

\st
It is then interesting to compare $\Pi^q_1$ with
a straightforward {\fASPcr} encoding, $\Pi^q_2$:
\[
\small
\begin{array}{l}
\hif Q_1 < Q_2, col(Q_1)=col(Q_2). \\
\hif Q_1 < Q_2, row(Q_1)=row(Q_2). \\
\hif Q_1 < Q_2, col(Q_2)-col(Q_1) = | row(Q_2)-row(Q_1) |.
\end{array}
\normalsize
\]
Condition $Q_1 < Q_2$ performs basic 
symmetry breaking. $\Pi^q_2$ uses two functions
to encode the positions of the queens.
\emph{What is remarkable about $\Pi^q_2$ is that the grounding
of the last rule grows roughly with $O(n^2)$, although
we argue that it is as straightforward an encoding of the
requirement as the corresponding rule from $\Pi^q_1$}.
As in the previous scenario, we expect a similar growth
for comparable \textsc{clingcon} encodings, and
a growth of $O(n^4)$ for the grounding of the encodings written in
the languages of \cite{cab11,lif11}.

%
%

\section{Conclusions and Future Work}
In this paper we have defined the syntax and
semantics of an extension of ASP 
by non-Herbrand
functions with full-fledged arithmetic expressions 
and consistency-restoring rules.
The resulting language {\fASPcr} supports
partial functions and we hope we have demonstrated
that it allows for the encoding
of rather sophisticated kinds of knowledge, including
knowledge about total functions.
Compared to similar languages, {\fASPcr}
strikes a remarkable balance between
expressive power and efficiency of 
computation.
In the previous section, the discussion on the 
efficiency of computation was based only on the
size of the grounding of the corresponding
encodings, but in \cite{bal12b} experimental
evidence on solver performance was obtained using a prototype of
an {\fASP} solver (available at {http://marcy.cjb.net/clingof}).
We expect that a version of the solver including
support for the extended language 
defined in this paper will be available soon.
Once that becomes available, we plan to substantiate
the discussion from the previous section with
experimental results.

\bibliographystyle{splncs03}
\bibliography{biblio}

\end{document}